\documentclass[11pt]{article}
\pdfoutput=1
\usepackage[utf8]{inputenc}
\usepackage{authblk}
\usepackage{hyperref}
\usepackage{tikz}
\usepackage{setspace}
\usepackage{natbib}
\usepackage[margin=1.25in]{geometry}
\usepackage{graphicx}
\graphicspath{ {./figures/} }
\usepackage{subcaption}
\usepackage{amsmath}
\usepackage[linguistics]{forest}

\title{A Study of Genetic Algorithms for Hyperparameter Optimization of Neural Networks in Machine Translation}

\author{Keshav Ganapathy}

\date{}

\begin{document}

\maketitle

\begin{abstract}
\noindent With neural networks having demonstrated their versatility and benefits, the need for their optimal performance is as prevalent as ever. A defining characteristic, hyperparameters, can greatly affect its performance. Thus engineers go through a process, tuning, to identify and implement optimal hyperparameters. That being said, excess amounts of manual effort are required for tuning network architectures, training configurations, and preprocessing settings such as Byte Pair Encoding (BPE). In this study, we propose an automatic tuning method modeled after Darwin's Survival of the Fittest Theory via a Genetic Algorithm (GA). Research results show that the proposed method, a GA, outperforms a random selection of hyperparameters.
\end{abstract}


\section{Introduction}
As neural networks are being adopted to solve real-world problems, while some parts of the network may be easy to develop, other unknown aspects such as hyperparameters, have no clear method of derivation. Ongoing research focuses on developing new network architectures and training methods. When developing neural networks, the question at hand is how to set the hyperparameter values to maximize results and set the training configuration. For network architecture design, important hyperparameters include the type of network, the number of layers, the number of units per layer, and unit type. For training configurations, important hyperparameters include learning algorithm, learning rate, and dropout ratio. All these hyperparameters interact with each other and affect the performance of neural networks. This interaction between hyperparameters can be referred to as epistasis. Thus they need to be tuned simultaneously to get optimum results.\\

The motivation behind this research is to replace tedious manual tuning of hyperparameters with an automatic method performed by computers. Current methods of optimization are limited to trivial methods like Grid search. Grid search is a simple method for hyperparameter optimization. However, as the number of hyperparameters increases, Grid search becomes time consuming and computationally taxing. This is because the number of lattice points increases in an exponential way with an increase in the number of hyperparameters \cite{qin2017evolution}. For example, if there are ten hyperparameters to be tuned and we only try five values for each parameter, and this alone requires more than 9 Million evaluations: \begin{math}5^{10} = 9765625\end{math}. For this reason, the grid search is not feasible for certain applications. To solve this, we look to a GA for a higher-performing and less computationally taxing solution. The use of a GA for neural network hyperparameter optimization has been explored previously in \cite{suganuma2017genetic, moriya2018evolution}. \\

We present an empirical study of GAs for neural network models in machine translation of natural language specifically Japanese to English. We describe the experiment setup in Section 2, our GA method in Section 3,  and results in Section 4. The preliminary findings suggest that a simple GA encoding has the potential to find optimum network architectures compared to a random search baseline.

\section{Experimental Setup}

Genetic Algorithms (GA) are a class of optimization methods where each individual, a neural network, represents a solution to the optimization problem and the population is evolved in hopes of generating good solutions. In our case, each individual represents the hyperparameters of a neural machine translation system, and our goal is to find hyperparameters that will lead to good systems. The defining factor when measuring an individual's fitness is its BLEU score, a measurement of the individual's translation quality which is dependent on the individuals hyperparameters. The data set used for experimentation was limited to 150 individuals each consisting of 6 hyperparameters. We use a benchmark data set provided by \cite{zhang2020benchmarks} where a grid of hyperparameter settings and the resulting model BLEU scores are pre-computed for the purpose of reproducible and efficient hyperparameter optimization experiments. In particular, we use the Japanese-to-English data set which consists of various kinds of Transformer models (described later) trained on the WMT2019 Ro-bust Task \cite{li-EtAl:2019:WMT1}, with BLEU scores ranging from 9 to 16.\vspace{6pt}

Every test was done over three trials each consisting of 1000 optimization iterations. One iteration involves a process, elaborated below, to arrive at the target BLEU score of 16. The higher the BLEU score the better. While testing, both the GA and the Baseline (random search) received the same initial population of 5, 10, 15, 20, or 25 individuals. Throughout experimentation, the goal was to arrive at one individual with a BLEU score, fitness, of 16 or higher. The fitness is a measure of the likelihood of the individual remaining in the population.\vspace{6pt}
 
A practical limitation of the data set includes a possibility that a combination of hyperparameters found by the GA may not be represented in the data set. If the combination is non-existent, we assign a fitness value of 0 to that respective individual. This is imperative as we do not want a non-existant individual remaining in the population. By assigning the individual a fitness value of 0, it is guaranteed be replaced in the next breeding cycle. Furthermore, the possibility of two individuals with a fitness 0 is impossible as the population starts with individuals with representation in the data set. This is a benefit of an individual based measurement rather than a generational measure. Additionally,  during experimentation, individuals were not added to the population if they have already been in the population, or if their genes exist in the current population. In traditional metrics, the entire current population is replaced by a new generation consisting of new offspring. In our implementation, only one offspring is generated, and that replaces the weakest individual. The performance of the GA and baseline algorithm is represented with a value based on the total amount of individuals added. This value increases every time an individual is added to the population. The performance of both algorithms is determined by averaging the performance value measured from all the iterations. An iteration consists of one optimization cycle. The performance value measured in individuals per iteration is displayed in the tables in section 4. 

\section{Method} 
The GA system used in this experiment is based on the natural selection process theorized by Darwin. This theory states that the stronger/fitter individuals survive while the weaker individuals do not. Thus over time, when the surviving stronger individuals reproduce, you get a population that as a whole carries genes that make them more resilient. However, Darwin only theorized this process in the natural world. The idea for using a GA in optimization problems and machine learning was thought of by Goldberg and Holland and expressed in \cite{goldberg1988genetic}. Throughout this experiment, we simulate their logic for machine translation neural network optimization. We begin with an initial population that "reproduces" until we get a group of individuals that collectively hold hyperparameters that perform better and one individual that is the "fittest" and meets the BLEU score target. The parts of the algorithm are represented through 3 objects: Individuals, Populations, and the Genetic Algorithm itself. The GA and Baseline algorithm were developed in the Python programming language. The GA and Baseline algorithm (random), elaborated later, are compared via an evaluator. This hierarchy, explained in \cite{whitley1994genetic}, and is represented visually below:\vspace{6pt}

\begin{figure}[ht]
    \begin{center}
    \begin{forest}
      [Evaluator
        [Genetic Algorithm
         [Population
            [Individuals]
         ]
        ]
        [Random Selection
         [Population
            [Individuals]
         ]
        ]
      ]
    \end{forest}
     \caption*{Figure 1: Hierarchy of the Objects in Experimentation}
    \end{center}
\end{figure}
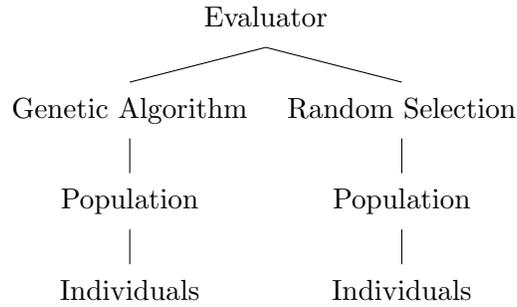

\subsection{Individual}
Every individual consists of two defining characteristics: 1) the "chromosome", consisting of 6 hyperparameters, and 2) the fitness score, which is the BLEU value. An individual's chromosome can look like [10000.0, 4.0, 512.0, 1024.0, 8.0, 0.0006]. The fitness of the machine translation neural networks (a.k.a individual) is defined as the BLEU score or how well the neural network can translate from Japanese to English. BLEU score is measured by an algorithm for evaluating the quality of translation performed by the machine to that performed by a human. The BLEU score is calculated via a simple proportion. The two translations, the human translation, referred to as reference translation, and the translation done by the machine, the candidate translation. To compute, one counts up the number of candidate translation words, unigrams, that occur in any reference translation and the total number of words found in the reference translation \cite{papineni2002bleu}. Once we get the two values, we divide them to get a precision component of the BLEU score.\vspace{6pt}
 
However, simple proportions are not the only thing used during calculation of BLEU score. Another property of BLEU score worth noting is the use of a brevity penalty, a penalty based on length. Since BLEU is a precision based method, the brevity penalty assures that a system does not only translate fragments of the test set of which it is confident, resulting in high precision \cite{koehn2004statistical}. Is has become common practice to include a word penalty component dependant on length of the phrase for translation. This is especially relevant for the BLEU score that harshly penalizes translation output that is too short. Finally, in this study, we use alternate notation for readability. A BLEU score of 0.289 is reported as a percent 28.9\%.\\\\\\\\

\begin{figure}[ht]
    \centering
    \begin{tabular}{ |l | c |}
    \hline
          \textbf{\# BPE subword units} (1k) & 10, 30, 50\\
          \textbf{\# encoder/decoder layers} & 2, 4\\
          \textbf{\# word embedding dimensions} & 256, 512, 1024 \\ 
          \textbf{\# hidden units}  & 1024, 2048 \\ \textbf{\# attention heads} & 8, 16\\ \textbf{initial learning rate} ($10^{-4}$) &3, 6, 10\\
         \hline
    \end{tabular}
    \caption*{Table 1: Hyperparameter search space for the Tranformer NMT systems}
    \label{tab:datasets_hyps}
\end{figure}

Amongst the 150 individuals there are 7 target BLEU scores above the 16 goal: 16.04, 16.09, 16.02, 16.41, 16.03, 16.13, and 16.21. Ranges of hyperparameter values, referred to as Genes, are as follows: Fitness(9.86 - 16.41), Gene 1 (10,000, 30,000, or 50,000), Gene 2 (2.0 or 4.0), Gene 3 (256.0, 512.0, or 1024.0), Gene 4 (1024.0 or 2048.0), Gene 5 (8.0 or 16.0), and Gene 6 (0.001, 0.0003, or 0.0006). This information is shown above.\\

Additionally, a convolutional or recurrent model was not implemented for these translation networks. The networks in this study implemented transformers. On top of higher translation quality, the transformers requires less computation to train and are a much better fit for modern machine learning hardware, speeding up the training process immensely. In regards to the less computational power needed, the ease of training of transformers can be accredited to its lack of growth based on amount of words \cite{dehghani2018universal}. Specifically, a transformer is not recurrent meaning it does not need the translation of the previous word to translate the next. For example lets take the example a translation for German to English. Let's use the German sentence, Das Haus ist groß, meaning the house is big. In a Recurrent Neural Network (RNN), the network identifies das and the, and then uses that as a reference for the next word to translate Haus to house, etc. However, in a Transformer, we can treat each word as a separate object and for translation rather than the translation of the previous word, it uses the embedding value of all of the other words. Because they are treated independently, we can have the translation operation occur in parallel. So in this example every word das, Haus, ist, groß are all vectored and use the other words' embeddings. Additionally, a notable characteristic of a transformer is its use of an attention mechanism. The Attention mechanism allows for the network to direct its focus, and it pays greater attention to certain factors when processing the data resulting in a higher performing network. For these three main reasons, the lack of recurrence, the use of an attention mechanism, and the ability for parallel computation, transformers are a preferred choice as a network architecture in Machine Translation. The hyperparameters searched for in our Transformer models are shown in Figure 2.
   
\subsection{Population}
    The population of individuals consist of three instance variables: the population itself, consisting of an initial x individuals implemented through an array, and two individuals that resemble the fittest and second fittest individuals in the population. For a population of 5 individuals it will look something like [Individual 1, Individual 2, Individual 3, Individual 4, Individual 5]. Examples of the fittest and second fittest individual follow: Fittest: Individual 1 can be represented as the fitness with BLEU score of 16.41, and second fittest: Individual 2 can be represented as second fitness with BLEU score of 16.04. For experimentation, however the algorithm stopped when the goal of 16 or above is reached, so the situation above would not occur. 

\subsection{Genetic Algorithm (GA)}
    In a broad sense, a genetic algorithm is any population-based model that uses various operators to generate new sample points. The GA system used during experimentation abides to most conventional characteristics of GA: a population, individuals, selection/mutation/crossover operations, etc. Our GA is comprised of three objects: the population, a list of all individuals, and an individual that acts as the child, referred as place holder. The population holds the individuals, the list of all individuals allows us to make sure that the child is not a repeated individual, and the individual allows for an object to store information on the offspring. The following describes the structure of the GA: First, an array of individuals with only the current population. The current population is defined as the population before the selection process, elaborated below. Second, it will contain a List of all the individuals that have been introduced to the population. We iterate through this list every time before adding an individual to make sure that the new individual, or placeholder, has not been introduced before. Place holder is an individual that is initially set to have values of 0: [0.0, 0.0, 0.0, 0.0, 0.0, 0.0]. During the crossover process, elaborated later, all the genes are changed to that of the offspring. Our implementation, however, differs from convention in two main ways: our implementation includes an Integer Representation rather than bit value, and an individual based measure for optimization rather than a generational measure. At the end of the process, a value that represents the total number of individuals added to the population added to the initial population size is returned.\vspace{6pt}
\begin{center}
    The process the Genetic Algorithm goes through is depicted below:
    \begin{tikzpicture}[node distance=1.5cm]
        \node (io) [rectangle] {Genetic Algorithm};
        \node (pro1) [rectangle, below of=io] {Selection};
        \node (dec1) [rectangle, below of=pro1] {Crossover};
        \node (pro2) [rectangle, below of=dec1] {Mutation};
        \node (dec2) [rectangle, below of=pro2] {Check Validity of Individual};
        \node (pro3) [rectangle, below of=dec2] {Add Individual};
        \node (dec3) [rectangle, right of=dec1, xshift=2cm,yshift=-.75cm] {Not Valid};
        \node (dec4) [rectangle, below of=pro3] {Check if target is reached};
        \node (pro4) [rectangle, below of=dec4] {Optimization Complete};
        \node (dec5) [rectangle, left of=pro2, xshift=-3cm,yshift=-.75cm] {Target not Reached};

        \draw [->] (io) -- (pro1);
        \draw [->] (pro1) -- (dec1);
        \draw [->] (dec1) -- (pro2);
        \draw [->] (pro2) -- (dec2);
        \draw [->] (dec2) -- node {Valid} (pro3);
        \draw [->] (dec4) -- node {Target Reached} (pro4);
        \draw [->] (dec3) |- (pro1);
        \draw [->] (dec5) |- (pro1);
        \draw [->] (dec3) |- (pro1);
        \draw [-] (dec4) -| (dec5);
        \draw [->] (pro3) -- (dec4);
        \draw [-] (dec2) -| (dec3);         
    \end{tikzpicture}
\\The processes used during the GA, selection, crossover, and mutation, and their functionalities are elaborated on below.
\end{center}
\subsubsection{Selection}
    The selection process allowed for the GA to select two parents to mate to form a new offspring. The two parents are selected by weighted probability, proportional to their fitness. For example, lets take an initial population of 5 individuals with fitness values 10, 25, 15, 5, 45. The probability of selecting an individual is calculated by finding the sum of the fitness values, in this example 100, divided by the individuals fitness. Individual 1 has a 10\% of being selected, 10/100, Individual 2 has a 25\% chance, 25/100,	etc. After repeating this process we get a list of percentages 10\%, 25\% , 15\% , 5\% , 45\%. As you can see, the percentages will always add up to 100. These values are then used to get an array with range values for each individual. In the aforementioned example, a list will look like [10, 35, 50, 55, 100]. These values are determined by adding the percentage value of an individual to that of all the previous individuals. From here, we select a random integer value from 0 to 100. This value is then correlated to an individual. For the above example with a population size of 5 Individual 1 is selected if the random value is less than 10, Individual 2 is selected if the value is greater than 10 and less than 35, Individual 3 if the value is greater than 50 and less than 55, etc. This approach is optimal as its adaptable based of size of population, and gives an accurate weighted representation.

\subsubsection{Crossover}
    The crossover process simulates the breeding part of the natural selection process. Following selection, the selected two individuals are used to create an offspring. Initially, a random gene in the chromosome is selected as the "crossover point". Up to that point genes of the fittest, Parent 1 in Figure 2, are selected, and from that point to the end, genes of the second fittest, Parent 2 in Figure 2, are added. For a cross over point of 2, the example below depicts the cross over process.
    \begin{figure}[ht]
        \begin{center}
            \includegraphics[width=1\textwidth]{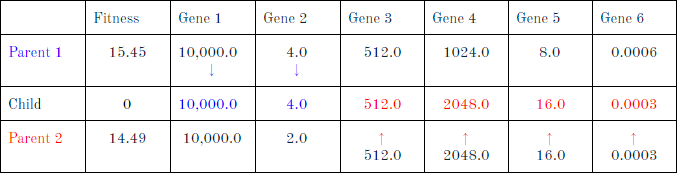}
            \caption*{Figure 2: Visual depicting the Crossover process}
        \end{center}
    \end{figure}
    \\ From here we now need to assign a fitness a value to the individual. Later we explain how we derived the 7 lists, for now just understand that we have 7 lists, one list with all potential values for each hyperparameter respectively, and another list for every fitness a value. These lists are ordered by individuals. For example the first index in all 7 lists correspond to individual one, the second index to individual 2, etc. We initially iterate through the entire first list, an array containing all values for the 150 individuals ordered, until we arrive at a match. In the example above this can be seen when we first arrive at the value of 10000. Then we store the index of the 10000 and see if that same index in list 2,3,4 etc also match the individuals gene. If all 6 genes correlate to values at one index, we assign a fitness value from the fitness list at the same index. Though an inefficient operator for assigning fitness, this approach makes the mutation operator easier. 

\subsubsection{Mutation}
    During the experiment, a mutation rate of 1/8, 12.5\%, was selected. When mutation occurs, an individual in the population is randomly selected, and one gene of that individual is also randomly selected. That gene is then given a new random value from the data set. Mutation process excludes the weakest Individual from being selected as that Individual would be replaced. If the weakest individual were not excluded, it would be replaced in during the addition of the offspring making the mutation irrelevant. Below is a mutation example with the mutation occurring on Gene 5. As shown, the mutation is resulting in an increased fitness. 
    \begin{figure}[ht]
      \begin{center}
        \includegraphics[width=1\textwidth]{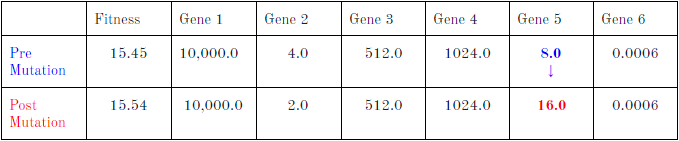}
        \caption*{Figure 3: Visual depicting the Mutation process}
    \end{center}
    \end{figure}

\subsection{Evaluator}
    All of the below results are measured using an average of individuals needed to be added over 1000 iterations. Each iteration consists of a process of going through optimization until a target score of 16 is reached. As previously stated, both algorithms return a value that represents the number of individuals that were added to reach the goal. However, before returning the values, we add the initial population size to account for individuals in the initial population. These values are then stored in two lists one for each algorithm, one list for the GA and another for the Baseline algorithm. At the very of the end of the evaluator the average of both lists is used to arrive at performance measure of the algorithms.
\subsection{Baseline (Comparison Algorithm)}
    As a measure of the quality of the GA, a baseline algorithm that optimizes by randomly selecting hyperparameters was used to compare. The baseline algorithm selected a random index value from 0 to one less than the size of the data set. From this, an individual is made with values from that index in every list from the initialization of data points, elaborated below. At the end of the process, a value that represents the total number of individuals added to the population added to the initial population size is returned.
    \begin{center}
    The process the Baseline Algorithm goes through is depicted below:
        \begin{tikzpicture}[node distance=1.5cm]
            \node (io) [rectangle] {Baseline Algorithm};
            \node (pro1) [rectangle, below of=io] {Select Random Value};
            \node (dec4) [rectangle, below of=pro1] {Get Individual that corresponds to the Random Value};
            \node (dec1) [rectangle, below of=dec4] {Check Validity of Individual};
            \node (pro2) [rectangle, below of=dec1] {Add Individual};
            \node (dec2) [rectangle, right of=dec4, xshift=4.5cm] {Not Valid};
            \node (pro3) [rectangle, below of=pro2] {Check if target is reached};
            \node (dec3) [rectangle, below of=pro3] {Optimization Complete};
            \node (pro4) [rectangle, left of=dec1, xshift=-5cm] {Target not Reached};

            \draw [->] (io) -- (pro1);
            \draw [->] (pro1) -- (dec4);
            \draw [->] (dec4) -- (dec1);
            \draw [->] (dec1) -- node {Valid} (pro2);
            \draw [->] (pro2) --  (pro3);
            \draw [->] (pro3) -- node {Target Reached} (dec3);
            \draw [-] (dec1) -| (dec2);
            \draw [->] (dec2) |- (pro1);
            \draw [->] (pro3) -| (pro4);
            \draw [->] (pro4) |- (pro1);
        \end{tikzpicture}
    \end{center}
\subsection{Initialization of Data Points}
    The initialization uses the python pandas library to create seven arrays that are representative of all possible Individual combinations. For example, all potential Gene 1 values are stored in an array. Similarly, there are arrays for the other genes and the fitness values. Additionally, these arrays store the fitnesses that are in order of fitness. Index 0 of all the lists correlate to the characteristics of individual 1, index 1 for individual 2, etc. These arrays are implemented to assign fitness values when a new off-springs are created, generate the initial population, and during the mutation operator.

\section{Results}
    Every trial in the tables below consists of 1000 iterations. An iteration entails the repetition of the GA and Baseline processes, elaborated above, until the target goal is reached. Thus the average is calculated over an accumulative 3000 (per the tables below) iterations. We measure how long it takes for an algorithm to find the a good solution (i.e. $\ge 16$ BLEU), so the lower the better.
\begin{center}
\end{center}
    \begin{figure}[ht]
        \begin{center}
        \begin{tabular}{ |p{2.5cm}||p{2.5cm}|p{2.5cm}|p{2.5cm}| p{2.5cm}|  }
            \hline
            \multicolumn{5}{|c|}{Genetic Algorithm's Results} \\
            \hline
            Initial Population Size & Trial 1 (Average Number of Individuals Added)& Trial 2 (Average Number of Individuals Added)&Trial 3 (Average Number of Individuals Added) & Average\\
            \hline
            5&20.056&19.91&20.247&20.071\\
            10&21.011&20.739&20.392&20.714\\
            15&23.91&23.933&23.54&23.794\\
            20&27.555&28.019&27.267&27.614\\
            25&31.475&30.816&32.099&31.463\\
            \hline
        \end{tabular}
        \end{center}
        \caption*{Table 2: Genetic Algorithm Results}
        \label{fig:1}
    \end{figure}
    \begin{figure}[ht]
        \begin{center}
            \begin{tabular}{ |p{2.5cm}||p{2.5cm}|p{2.5cm}|p{2.5cm}| p{2.5cm}| }
            \hline
            \multicolumn{5}{|c|}{Baseline Algorithm Results} \\
            \hline
            Initial Population Size & Trial 1 (Average Number of Individuals Added)& Trial 2 (Average Number of Individuals Added)&Trial 3 (Average Number of Individuals Added)&Average\\
            \hline
                5&20.435&21.949&21.891&21.425\\
                10&23.513&22.545&23.29&23.116\\
                15&26.056&25.611&25.345&25.671\\
                20&28.332&28.292&28.365&28.330\\
                25&31.857&30.819&32.117&31.598\\
            \hline
            \end{tabular}
        \end{center}
        \caption*{Table 3: Baseline Algorithm Results}
        \label{fig:2}
    \end{figure}

    \begin{figure}[h]
        \begin{center}
        \vspace{1em}
        \vspace{1em}
    \begin{tabular}{|p{4cm}||p{4cm}||p{4cm}|}
    \hline
    \multicolumn{3}{|c|}{Performance Difference} \\
 \hline
 Initial Population Size & Winner & Difference in Performance\\
 \hline
 5 & Genetic Algorithm & 1.354\\
 10 & Genetic Algorithm & 2.402\\
 15 & Genetic Algorithm & 1.877\\
 20 & Genetic Algorithm & .716\\
 25 & Genetic Algorithm & .135\\
 \hline
\end{tabular}
\end{center}
        \caption*{Table 4: Difference in Performance between GA and Baseline Algorithm:}
        \label{fig:3}
\end{figure}

\vspace{144pt}Using the average values, shown below, we can get a numerical representation of how much better a GA is. The average of the values showing difference in the performance shows that the GA can reach the desired goal with an average of 1.27 individuals fewer. The values in the table above were found by taking the average from the Baseline and subtracting the correlating GA value to find how many more individuals the Baseline needs on average. The value 1.27 was found by averaging all the values in the table below. Although saving 1.27 iterations is not large in the grander scheme of things, it is promising to see that GA gives consistent gains, implying that there are patterns to be exploited in the hyperparameter optimization process.\\\\\\\\\\

\section{Future Work}
The study proves validity of a GA being implemented for hyperparameter optimization. Future work falls into three main categories: general, structural (changes how the system works), and behavioral changes (variations in individual methods implementations which can cause varying results). Additionally, there is the possibility for variations to be implemented for the Baseline algorithm. Lastly, we will explore results of the generational measure rather than the individual based. We realized that a plateua occurs as the population increases. This is due to the chance of picking individuals with lower fitness values increasing as the population increases for the GA. Therefore, as the initial population size increases, the GA's performance approaches that of the Baseline algorithm.

\subsection{General Changes}
When reflecting on the experiment as a whole, there are many key aspects that should be changed to further test the validity of a GA. One such example includes the range of Individuals represented in the data set. For example, in the data set above, many combinations of hyperparameters where not pre-trained so a fitness value was not assigned. This results in the GA iterating extra times due to non-existant individuals. Additionally, as previously mentioned, one goal of a GA is to outperform the grid search, a more primitive type of optimization. To further test the performance, a larger data set will need to be used to test the efficiency compared to grid search, a higher performing solution than random. Also, additional evaluation metrics can be used such as the populations average fitness and run time of the algorithm. Another potential change can be discounting individuals that are not represented. While impractical to have all combinations represented, the Baseline algorithm could not select a combination that was not represented while the GA could. For further comparison, not adding an individual to the population would decrease the individual count and enlarge the performance gap between the GA and Baseline Algorithm.

\subsection{Structural Changes}
An example of a structural change includes: Having mutation occur before selection. The mutation before selection can result in different weights of individuals during the selection process. This can result in varying fitness values which can ultimately change the entire optimization process.

\subsection{Behavioral Changes}
A glaring example of a behavioral change can be seen in how the individual that is being replaced is selected. Unlike Darwin's theory, the replaced individual was selected definitively during the experiment. The individual with the lowest fitness was replaced. Changing this to weighted probability similar to selecting the fittest and second fittest can affect the tuning process as hyperparameters work simultaneously. Two individuals with low fitness values can have a child that has a high fitness value due to the lack of direct correlation between individual hyperparameter values and BLEU score. Another example includes a two-point crossover operator rather than the one-point crossover currently implemented. A two-point cross over has the individuals exchange the genes that fall between these two points.

\section{Conclusion}
This work introduces an advanced GA for hyperparameter optimization and applies it to machine translation optimization. We demonstrate that optimization of hyperparameters via a GA can outperform a random selection of hyperparameters. Specifically, outperform is defined by the ability of the algorithm to arrive at the goal with less individuals added. Finally, we propose future research directions which are expected to provide additional gains in the efficacy of GAs.

\section{Acknowledgements}
I would like to thank and acknowledge Dr. Kevin Duh (JHU) for giving me the opportunity to pursue research with him. Thank you for the continuous support, patience, and motivation throughout the entire process. Additional thanks for the feedback and comments on this paper, and for the invaluable guidance regarding programming, resources for help, and much more.\\

Also, I am very appreciative of and grateful to my family and friends for their love, patience and support, without which this project would not have been possible.

\bibliography{main}

\end{document}